\documentclass[conference]{IEEEtran}
\IEEEoverridecommandlockouts
\usepackage{cite}
\usepackage{amsmath,amssymb,amsfonts}
\usepackage{algorithmic}
\usepackage{graphicx}
\usepackage{textcomp}
\usepackage{xcolor}
\usepackage{booktabs}
\usepackage{multirow}
\usepackage{makecell} 
\usepackage{tabularx}
\usepackage{subcaption} 
\usepackage{pifont}   %

\newcolumntype{C}{>{\centering\arraybackslash}X}

\def\BibTeX{{\rm B\kern-.05em{\sc i\kern-.025em b}\kern-.08em
    T\kern-.1667em\lower.7ex\hbox{E}\kern-.125emX}}
\begin{document}

\title{Robust Embodied Perception in Dynamic Environments via Disentangled Weight Fusion }

\author{Juncen Guo\textsuperscript{1}, Xiaoguang Zhu\textsuperscript{2}, Jingyi Wu\textsuperscript{1}, Jingyu Zhang\textsuperscript{1}, Jingnan Cai\textsuperscript{1}, Zhenghao Niu\textsuperscript{1}, Liang Song\textsuperscript{1}$^\ast$\thanks{$^\ast$Corresponding authors. This work was supported by National Key Research and Development Program of China, Project No.2024YFE0200700, Subject No.2024YFE0200703. This work was also supported in part by the Specific Research Fund of the Innovation Platform for Academicians of Hainan Province under Grant YSPTZX202314, in part by the Shanghai Key Research Laboratory of NSAI and China FAW Joint Development Project.(Email: guojc23@m.fudan.edu.cn, songl@fudan.edu.cn) 
}
\\
\textsuperscript{1}Fudan University %
\\
\textsuperscript{2}University of California, Davis  %
}

\maketitle

\begin{abstract}

Embodied perception systems face severe challenges of dynamic environment distribution drift when they continuously interact in open physical spaces. However, the existing domain incremental awareness methods often rely on the domain id obtained in advance during the testing phase, which limits their practicability in unknown interaction scenarios. At the same time, the model often overfits to the context-specific perceptual noise, which leads to insufficient generalization ability and catastrophic forgetting. To address these limitations, we propose a domain-id and exemplar-free incremental learning framework for embodied multimedia systems, which aims to achieve robust continuous environment adaptation. This method designs a disentangled representation mechanism to remove non-essential environmental style interference, and guide the model to focus on extracting semantic intrinsic features shared across scenes, thereby eliminating perceptual uncertainty and improving generalization. We further use the weight fusion strategy to dynamically integrate the old and new environment knowledge in the parameter space, so as to ensure that the model adapts to the new distribution without storing historical data and maximally retains the discrimination ability of the old environment. Extensive experiments on multiple standard benchmark datasets show that the proposed method significantly reduces catastrophic forgetting in a completely exemplar-free and domain-id free setting, and its accuracy is better than the existing state-of-the-art methods.
\end{abstract}


\section{Introduction}

In recent years, with the deep integration of embodied intelligence and multimedia technology, interactive multimedia systems have been widely used in the fields of autonomous driving, intelligent robots and environmental monitoring\cite{p23}. Different from static vision tasks with fixed distribution of training data, embodied agents need to perceive and interact continuously in an open and dynamic physical environment\cite{p10}. In this process, the sensed data often arrive continuously in the form of streams, accompanied by significant shifts in the environmental domain such as illumination changes, weather changes, or sensor noise\cite{p23,p11}. Therefore, Domain Incremental Learning (DIL) is necessary to be applied to the embodied perception model to adapt to the new environment distribution while maintaining the memory ability of the old environment knowledge. This is of great research significance for building intelligent systems with long-term autonomy.

Within the category of continual learning, it is usually divided into task incremental, class incremental and domain incremental learning\cite{p22}. Compared with the scenarios with clear task boundaries, DIL for embodied perception faces a more severe challenge: when the agent performs the same type of task, it needs to deal with the sharp and non-stationary drift of the sensor input distribution\cite{p29}. Although catastrophic forgetting is the core problem, for resource-constrained mobile multimedia platforms or privacy-sensitive home service scenarios, traditional experience replay methods need to store a large amount of historical sensor data, which is no longer applicable\cite{p27}. Although the existing regularization and distillation methods can reduce the storage overhead, they are often difficult to balance between stability and plasticity when dealing with complex physical environment changes\cite{p1}.

\begin{figure}[t]
    \centering
    \begin{minipage}{0.41\linewidth}
        \centering
        \includegraphics[width=\linewidth]{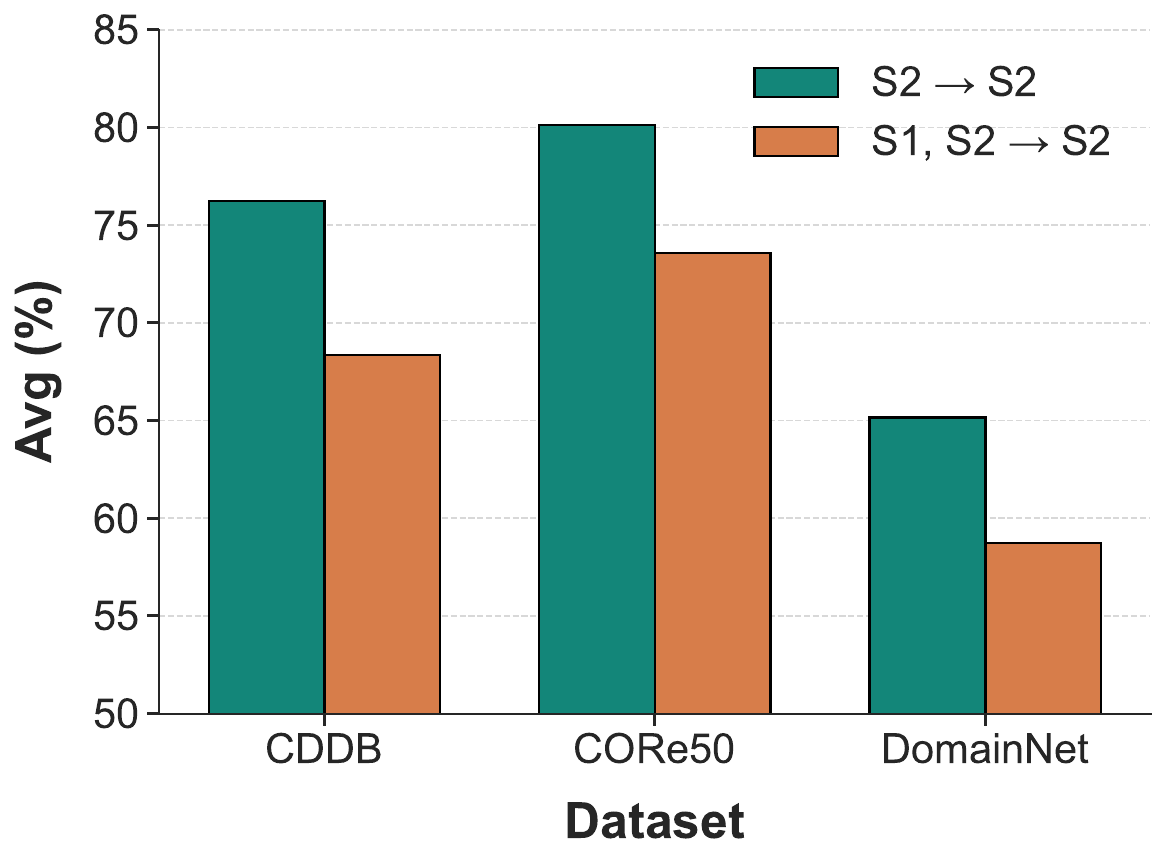}
    \end{minipage}
    \hfill
    \begin{minipage}{0.54\linewidth}
        \centering
        \includegraphics[width=\linewidth]{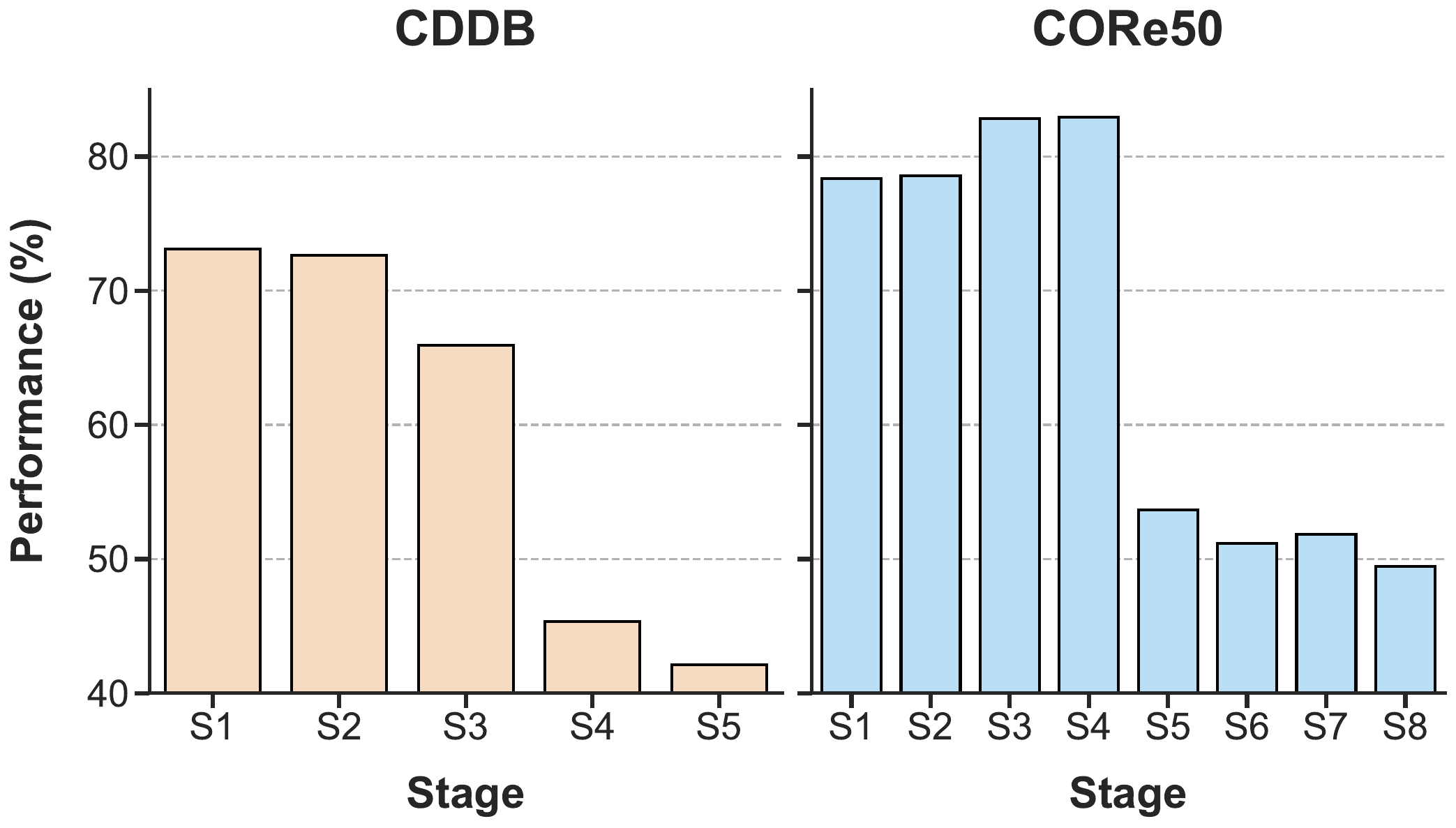}
    \end{minipage}
    \caption{The performance of method S-iPrompts\cite{p12} under different experimental conditions. \textit{Left}: the test performance of the model on domain S2 after training only on S2, and after training on both S1 and S2 (covering three datasets). \textit{Right}: the model undergoes training with S1-S3 and is then tested in the subsequent unseen domain with zero samples on the CDDB\cite{p3} dataset, the model undergoes training with S1-S4 and is then tested in the subsequent unseen domain on the CORe50\cite{p4} dataset.}
\vspace{-15pt} 
\label{fig1}
\end{figure}

Although the existing DIL research has made some progress, it still faces many limitations when constructing a robust embodied perception system\cite{p28}. Most existing DIL methods that based on prompt-based learning or specific domain components usually need to pre-match the domain id in the testing phase, or predict the domain of the sample through an additional module\cite{p14}. So that the corresponding parameter branch can be called. This approach essentially reduces the more complex DIL problem to the relatively easy TIL problem. However, in real-world scenarios, when an agent travels in an unknown physical space, it cannot predict the environment label. This dependence on prior information seriously limits the real-time interaction ability of the system in the open world\cite{p24}.
 And existing methods do not generalize well on unseen domains\cite{p12} as shown in Fig.~\ref{fig1}(\textit{Right}). Due to excessive attention to the specific distribution of the current domain, the model often tends to learn domain-specific noise information such as background texture, color style, rather than the intrinsic semantic features of the class itself. 
This sensitivity to non-essential environmental factors leads to a lack of robustness of the system in the face of sensor failures or abrupt changes in the environment. Additionally, although existing incremental strategies alleviate forgetting to some extent\cite{p9}, they often introduce the interference of the old domain knowledge on the discrimination of the new domain as shown in Fig.~\ref{fig1}(\textit{Left}). It is clearly evident that the learning process in domain S1 has severely impacted the performance of the model. The feature distribution of the old domain may form obstacles in the learning process of the new domain, resulting in a fuzzy discriminant boundary between the old and new domains. 
It is difficult to achieve a "win-win" feature representation within the limited parameter space of edge computing devices\cite{p30}.

\begin{figure*}[t]
   \centering
   \includegraphics[width=1\textwidth]{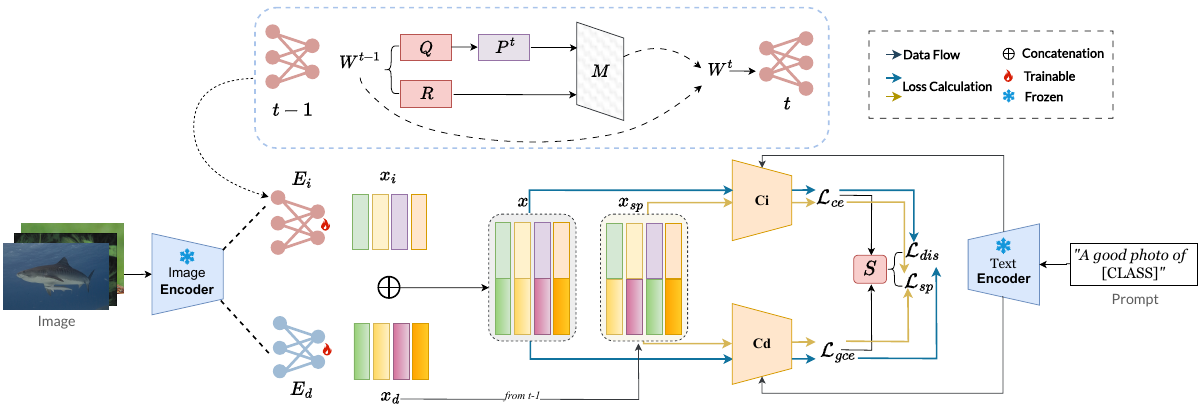}
   \caption{The framework of disentangled representation and weight fusion for DIL. %
   }
   \label{fig2}
   \vspace{-15pt}

   \end{figure*}

To address the above issues, this paper proposes a new DIL method without domain id matching and without experience replay. This method aims to break the explicit dependence of traditional DIL on domain information, and achieve efficient incremental learning by disentangling representation and weight fusion mechanism. Specifically, we design a disentangledd representation mechanism to effectively strip domain-specific style noise and guide the model to focus on extracting cross-domain shared category intrinsic features, thereby eliminating the interference of domain information on category discrimination. In order to mitigate catastrophic forgetting without storing old samples, we use the weight fusion strategy to dynamically integrate the old and new knowledge in the parameter space, so as to ensure that the model retains the discriminative power of the old domain to the greatest extent while absorbing the distribution of the new domain.
The main contributions of this paper are summarized as follows:
\begin{itemize}
\item We construct a perception DIL framework that does not require domain id and does not rely on explicit domain prediction in the testing phase, which truly solves the difficult problem of blind adaptation of embodied agents in unknown environments.
\item Through innovative disentangled representation learning, we successfully eliminated the interference of environmental style changes on semantic discrimination, significantly enhancing the generalization ability of the multimedia system in unseen scenarios and the efficiency of extracting intrinsic features.
\item Experimental results on multiple standard DIL benchmark datasets show that the proposed method significantly reduces catastrophic forgetting without replaying any old samples, and outperforms the existing state-of-the-art methods in accuracy.
\end{itemize}

\section{Methodology}
\label{sec3}
To address the challenges of catastrophic forgetting and prediction interference induced by domain shift in DIL, we proposes a novel continual learning framework based on disentangled feature augmentation in Fig.~\ref{fig2}. The core rationale of our method lies in explicitly disentangling domain-invariant intrinsic features from task-dynamic domain features. Furthermore, by leveraging counterfactual reasoning to synthesize virtual cross-domain samples, the framework facilitates the acquisition of robust domain-invariant representations.

\subsection{Disentangled Feature Representation Learning}
\label{sec}
In the DIL, an input image typically comprises two distinct types of information: intrinsic features that define the class, and domain features governed by the current task environment. To effectively disentangle these components, we design a two-stream encoder architecture. Given the general visual feature $\mathbf{h}$ extracted by a pre-trained CLIP model, we employ an intrinsic encoder $E_i$ and a domain encoder $E_d$ to project $\mathbf{h}$ into independent latent spaces, yielding the intrinsic feature $\mathbf{x}_i$ and the domain feature $\mathbf{x}_d$, respectively.

To structurally ensure feature orthogonality, we employ a cross-gradient blocking mechanism that constrains the intrinsic encoder $E_i$ to be optimized exclusively via the intrinsic classifier $C_i$, and the domain encoder $E_d$ exclusively via the domain classifier $C_d$. Building on this, we introduce a re-weighting strategy based on a relative difficulty score\cite{p1} to address the scarcity of domain-conflicting samples. The core motivation is that within a specific domain, domain features often exhibit statistical correlations with class labels. While such correlations aid in-domain classification, they become spurious and detrimental during cross-domain transfer. Consequently, we guide the domain branch $(E_d, C_d)$ to capture these domain-specific environmental distributions. We employ the generalized cross entropy (gce)\cite{p2} loss for the domain branch, leveraging its robustness against noisy labels to focus the model on learning dominant environmental statistical regularities. In contrast, the intrinsic branch $(E_i, C_i)$ utilizes the standard cross entropy (ce) loss to establish precise class decision boundaries.

We compute the relative difficulty weight $S(\mathbf{x})$ of each sample based on the loss discrepancy between the two branches:

\begin{equation}
   S(\mathbf{x}) = \frac{\mathcal{L}_{\text{ce}}(C_d(\mathbf{x}), y)}{\mathcal{L}_{\text{ce}}(C_i(\mathbf{x}), y) + \mathcal{L}_{\text{ce}}(C_d(\mathbf{x}), y)},
   \end{equation}

The weight $S(\mathbf{x})$ quantifies the reliance of the prediction on domain environmental information. When the domain branch achieves a low loss (indicating easy classification), the sample is identified as domain-aligned, implying a strong statistical correlation between the class label and the domain environment. Consequently, $S(\mathbf{x})$ is assigned a smaller value. Conversely, a high loss in the domain branch suggests that the label cannot be easily inferred from the background or environmental context and primarily depends on the object's intrinsic features, resulting in a larger $S(\mathbf{x})$. By incorporating this weight, the disentanglement loss $\mathcal{L}_{\text{dis}}$ compels the intrinsic encoder to prioritize samples where domain information is insufficient for classification, thereby facilitating the extraction of more essential, class-discriminative features.

\begin{equation}
   \mathcal{L}_{\text{dis}} = S(\mathbf{x}) \cdot \mathcal{L}_{\text{ce}}(C_i(\mathbf{x}), y) + \mathcal{L}_{\text{gce}}(C_d(\mathbf{x}), y),
   \end{equation}
where $\mathcal{L} _ { gce }   ( C _ { d } ( \mathbf{x} ) , y )=\frac { 1 - C _ { d } ^ { y } ( \mathbf{x} ) ^ { q } } { q }$,  $C _ { d } ( \mathbf{x} )$ and  $C _ { d } ^ { y } ( \mathbf{x} )$ are the softmax output of the domian classifier and its probability of belonging to the target class y, respectively. And $ q \in ( 0 , 1 ]$ is a hyperparameter that controls how much bias is amplified.

\subsection{Anti-forgetting via Intrinsic Weight Fusion}
In the continual process of DIL, models confront the classic stability-plasticity dilemma. On one hand, the model requires plasticity to adapt to the distributional shifts of the new domain $t$. On the other hand, it must maintain stability to prevent new information from overwriting the critical knowledge structures of the previous domain $t-1$ through drastic parameter updates. We observe that domain features are inherently non-stationary and environment-dependent, fluctuating drastically across tasks and thus lacking value for cross-task inheritance. Conversely, intrinsic features capture the physical essence of classes, and their feature space should maintain topological consistency across different domains. Therefore, we propose a weight fusion strategy exclusively for the intrinsic encoder $E_i$, aiming to achieve explicit knowledge inheritance at the physical parameter level.

To achieve fine-grained, on-demand inheritance, we propose an adaptive weight fusion mechanism based on QR decomposition. This method employs orthogonal decomposition to identify the critical structural subspace within the weights of the previous task. It essentially freezes the parameter projections within this subspace, while permitting the acquisition of new knowledge in the orthogonal complement space. Specifically, consider a weight matrix $\mathbf{W}^{t-1}$ from a layer of the intrinsic encoder at the conclusion of task $t-1$. We first apply QR decomposition to $\mathbf{W}^{t-1}$ to extract the orthogonal basis and the corresponding structural coefficients representing the prior knowledgez:

\begin{equation}
   \mathbf{W}^{t-1} = \mathbf{QR},
   \end{equation}

where $\mathbf{Q}$ denotes an orthogonal matrix, whose column vectors constitute an orthonormal basis for the feature space of the previous task. $\mathbf{R}$ is an upper triangular matrix that encodes the projection coefficients of the prior weights onto these basis vectors, thereby representing the structural information of the accumulated knowledge.

\begin{table}
\centering
\caption{Results on CDDB.}
\label{tab1}
\renewcommand{\arraystretch}{1.15}
\begin{tabularx}{\columnwidth}{l c C C}
\toprule
\textbf{Method} & \textbf{Buffer} & \textbf{Avg} & \textbf{Last} \\
\midrule
LRCIL\cite{p16}        & \multirow{3}{*}{50/class}  & 74.01 & 73.42 \\
iCaRL\cite{p6}        &                             & 73.98 & 71.86 \\
LUCIR\cite{p15}        &                             & 80.77 & 78.25 \\
\midrule
LRCIL\cite{p16}        & \multirow{3}{*}{100/class} & 76.39 & 75.99 \\
iCaRL\cite{p6}        &                             & 79.76 & 78.51 \\
LUCIR\cite{p15}        &                             & 82.53 & 80.68 \\
\midrule
SimpleCIL\cite{p8}    & \multirow{7}{*}{0/class}   & 60.80 & 63.40 \\
L2P\cite{p9}          &                            & 67.33 & 64.45 \\
DualPrompt\cite{p10}   &                            & 68.33 & 71.41 \\
CODA-Prompt\cite{p11}  &                            & 69.19 & 74.18 \\
S-iPrompt\cite{p12}    &                            & 68.51 & 72.76 \\
DUCT\cite{p13}         &                            & 84.14 & 85.10 \\
PINA\cite{p14}         &                            & 85.71 & 86.32 \\
\midrule
\textbf{Ours} & \textbf{0/class} & \textbf{86.35} & \textbf{87.12} \\
\bottomrule
\end{tabularx}
\vspace{-20pt} 
\end{table}

At the initialization stage of task $t$, we project the randomly initialized weights of the new task, $\mathbf{W}^{t}_{\text{init}}$, onto the orthogonal basis $\mathbf{Q}$ of the previous task. This projection allows us to analyze the distribution of the new weights within the established knowledge space. The projection component $\mathbf{P}^{t}$ is calculated as $\mathbf{P}^{t} = \mathbf{Q}^{\top} \mathbf{W}^{t}_{\text{init}}$. To quantify the compatibility between the new initialization and the prior knowledge structure, we evaluate the discrepancy between the projection $\mathbf{P}^{t}$ and the old structural matrix $\mathbf{R}$. This discrepancy reflects the extent to which the new weights might perturb the existing knowledge structure. Based on this, we generate an adaptive fusion mask $\mathbf{M}$ to regulate the degree of freedom for updates at each parameter unit:

\begin{equation}
   \mathbf{M} = \text{clamp} \left( \frac { \| P^{t} - R \| _ { F } } { \| R \| _ { \max } } + \beta , 0 , 1 \right),
   \end{equation}

where $| | \cdot | | _F$ denotes the frobenius norm and $\beta$ is a bias term. As $\mathbf{M} \to 0$, the parameter is deemed critical for structural integrity, necessitating the preservation of $\mathbf{W}^{t-1}$. Conversely, as $\mathbf{M} \to 1$, the direction implies redundancy, allowing the model to adapt via $\mathbf{W}^{t}_{\text{init}}$.

Finally, we perform an element-wise adaptive fusion of the old and new weights using the mask $\mathbf{M}$ to obtain the weights for task $t$:
\begin{equation}
   \mathbf{W}^{t} = (1 - \mathbf{M}) \odot \mathbf{W}^{t-1} + \mathbf{M} \odot \mathbf{W}^{t}_{\text{init}},
   \end{equation}

Through QR decomposition based adaptive fusion, the model automatically distinguishes between structural components requiring preservation and spatial regions allowing for reshaping at the subspace level. This mechanism achieves a dynamic balance between mitigating catastrophic forgetting and adapting to new domains without the need to store old exemplars.

\subsection{Domain-invariant Feature Augmentation}
Although the aforementioned disentanglement mechanism structurally separates the two types of features, the data distribution within a single task is often homogenous. Consequently, intrinsic features may still inadvertently overfit the current domain environment, forming spurious statistical correlations. To break these potential dependencies and enhance robustness against domain shifts, we draw inspiration from counterfactual reasoning\cite{p21} and propose a feature swapping based data augmentation strategy. This approach aims to synthesize diverse cross-domain samples within the feature space, simulating the placement of current objects into different domain environments. Specifically, within a training mini-batch, we concatenate the intrinsic feature $\mathbf{x}_i$ of sample $k$ with the domain feature $\tilde{\mathbf{x}}_d$ of a randomly selected sample $j$ from last domain to construct a counterfactual sample $\mathbf{x}_{\text{sp}} = [\mathbf{x}_i; \tilde{\mathbf{x}}_d]$. Since $\tilde{\mathbf{x}}_d$ originates from a different image, it possesses no natural statistical correlation with $\mathbf{x}_i$. This operation effectively constructs diverse domain combinations at the feature level, compelling the model to eliminate non-causal environmental interference.

To effectively optimize using these synthetic samples, we introduce the swap loss $\mathcal{L}_{\text{sp}}$, with the core objective of establishing the domain invariance of intrinsic features. This loss requires the intrinsic classifier $C_i$ to correctly predict the original label $y$ based on $\mathbf{x}_i$, even when the domain feature has been tampered, while the domain classifier $C_d$ must correctly identify the label $\tilde{y}$ corresponding to the new domain feature $\tilde{\mathbf{x}}_d$. The loss function is defined as:
\begin{equation}
   \mathcal{L}_{\text{sp}} = S(\mathbf{x}) \cdot \mathcal{L}_{\text{ce}}(C_i(\mathbf{x}_{\text{sp}}), y) + \mathcal{L}_{\text{gce}}(C_d(\mathbf{x}_{\text{sp}}), \tilde{y}).
   \end{equation}

By incorporating the relative difficulty weight $S(\mathbf{x})$ and generalized cross entropy, we further reinforce the model's focus on hard samples. Mathematically, this constraint forces the intrinsic encoder to remain insensitive to variations in $\mathbf{x}_d$, thereby focusing on extracting genuinely domain-invariant features and severing the spurious co-occurrence between specific domain environments and class labels.
The final overall optimization objective:is described as:
\begin{equation}
   \mathcal{L}_{\text{total}} = \mathcal{L}_{\text{dis}} + \lambda\mathcal{L}_{\text{sp}}.
   \end{equation}
where $\lambda$ is a hyperparameter used to adjust the importance of the feature augmentation. %

\section{Experiments}
\subsection{Experimental Setup}
\subsubsection{Datasets}
We evaluated our method on three benchmark datasets with distinct characteristics. CDDB\cite{p3} is targeted at continual deepfake detection tasks. For this dataset, we adopted the most challenging Hard Track, which encompasses five sequential deepfake domains and comprises approximately 27,000 images. CORe50\cite{p4} is an object recognition dataset consisting of 11 domains and 50 classess. Following standard protocols, we utilized 8 domains for incremental training, while the remaining 3 unseen domains were reserved for testing. DomainNet\cite{p5} serves as a large-scale DIL dataset featuring 6 domains with significant domain shifts and 345 classes. Its training and test sets contain over 400,000 and 170,000 images, respectively.

\subsubsection{Evaluation Metrics}
At the $t$-th incremental stage, the incremental accuracy is measured as the classification accuracy on all seen domains. To fairly compare the overall performance of different methods, we define the average incremental accuracy as the mean of the accuracies obtained across all learning stages:$Avg = \frac{1}{T} \sum_{i=1}^{T} A_i$, where $T$ denotes the total number of incremental stages, and $A_i$ represents the accuracy at stage $i$. Additionally, we report the $Last$, which is the average accuracy achieved after the final task.

\subsubsection{Implementation Details}
All experiments were conducted on an NVIDIA RTX 4090 GPU using PyTorch. We utilized the pre-trained CLIP (ViT-B/16) as the backbone. The intrinsic and domain encoders are implemented as 3-layer MLPs with a dropout rate of 0.1. A warm-up strategy is applied where the feature swapping mechanism activates after 1,000 training steps. The text prompt for CLIP follows the template "a good photo of [CLS]," where [CLS] denotes the specific class name.%

\begin{table}[]
\centering
\caption{Results on CoRe50.}
\label{tab2}
\renewcommand{\arraystretch}{1.15}
\begin{tabularx}{\columnwidth}{l c C C}
\toprule
\textbf{Method} & \textbf{Buffer} & \textbf{Avg} & \textbf{Last} \\
\midrule
ER\cite{p17}        & \multirow{5}{*}{50/class} & 80.10 & 82.06 \\
L2P\cite{p9}       &                           & 81.07 & 82.72 \\
BiC\cite{p15}    &                           & 79.28 & 80.97 \\
DyTox\cite{p18}     &                           & 79.21 & 81.23 \\
DER++\cite{p19}     &                           & 79.70 & 82.01 \\
\midrule
SimpleCIL\cite{p8} & \multirow{8}{*}{0/class}  & 70.92 & 74.80 \\
L2P \cite{p9}       &                           & 83.57 & 87.87 \\
DualPrompt\cite{p10} &                          & 84.53 & 87.27 \\
CODA-Prompt\cite{p11} &                         & 87.92 & 91.57 \\
S-iPrompt\cite{p12} &                           & 81.95 & 83.38 \\
DUCT\cite{p13}      &                           & 91.95 & \textbf{94.47} \\
PINA\cite{p14}      &                           & 87.38 & 89.82 \\
\midrule
\textbf{Ours} & \textbf{0/class} & \textbf{92.36} & 93.57 \\
\bottomrule
\end{tabularx}
\vspace{-20pt} 
\end{table}

\subsection{Comparison with Existing Methods}
We comprehensively compare our method with the state-ofthe-art domain incremental learning method. These comparison baselines are mainly divided into two categories: one is the traditional exemplar-based method, and the other is the exemplar-free prompt-based method. In order to ensure the fairness of comparison, all methods uniformly use the pre-trained CLIP image encoder and text encoder as the backbone network
\subsubsection{Results on Dataset CDDB}
Table~\ref{tab1} shows the quantitative comparison results of the proposed method and the current two categories of state-ofthe-art DIL methods on the CDDB dataset. Overall, our method achieves an average accuracy of 86.35\%, and reaches 87.12\% on the last metric, outperforming all comparison baselines. This strongly proves that learning intrinsic features by disentangled representation is still an effective way to improve the generalization ability of the model even in complex scenes with large domain differences. Specifically, compared with the best performance of traditional case-based methods such as LRCIL, iCaRL and LUCIR, our method achieves 9.96\%, 6.59\% and 3.82\% performance improvement respectively without storing the old domain samples at all. This method not only consolidates knowledge in the parameter space through the weight fusion mechanism, but also avoids the risk of data leakage and storage overhead in privacy sensitive scenarios, and the final accuracy is significantly better than that of the strategy relying on replay. Furthermore, compared with the existing case-based prompt learning methods such as L2P and PINA, our method demonstrates clear superiority. %
\subsubsection{Results on Dataset CORe50}
Table~\ref{tab2} shows the performance of the proposed method on the Core50 dataset. Unlike CDDB, Core50 mainly examines the incremental adaptation ability of the model in the face of continuously changing background, illumination, and shooting viewpoint, and is a key benchmark to evaluate the robustness of real-world object recognition. Experimental results show that our method achieves an average accuracy of 92.36 on Core50, which significantly outperforms the existing exemplar-free baseline methods. This is mainly attributed to the fact that we effectively address the challenges posed by the specific background environment in Core50. Existing prompt-based methods tend to capture domain-specific context information, which is easy to cause the model to misjudge the background texture as a class feature. In contrast, our method successfully separates object semantic features from environmental background noise by using disentangled representation mechanism. Even without playing back any old samples, the accuracy is still 0.41\% higher than that of the suboptimal method DUCT, which strongly proves that our model can effectively ignore background interference and focus on the essential features of the object itself. In addition, for the more serious catastrophic forgetting problem in the longer task sequence of Core50, thanks to the weight fusion strategy, our method realizes the smooth integration of new and old knowledge in the parameter space. Compared with the ER and other methods that rely on old samples, our method not only completely eliminates the storage overhead, but also keeps ahead in the accuracy of the final stage. %
\subsubsection{Results on Dataset DomainNet}
Table~\ref{tab3} shows the performance of the proposed method on DomainNet large benchmark dataset. Experimental results show that our method achieves an average accuracy of 71.12\% and 69.97\% on the last metric, significantly outperforming all the compared exemplar-free baseline methods. Existing prompt learning methods, such as L2P and DUCT, tend to overfit the artistic style of a specific domain, while ignoring the structural semantics of the object itself. By disentangling the representation mechanism, our method forces the model to disentangle non-essential artistic style information from object semantic content, thereby learning a robust representation shared across styles, outperforming the second-best model by 2.06\% even when we do not store any exemplarity. %

\begin{table}[t]
\centering
\caption{Results on DomainNet.}
\label{tab3}
\renewcommand{\arraystretch}{1.15}
\begin{tabularx}{\columnwidth}{l c C C}
\toprule
\textbf{Method} & \textbf{Buffer} & \textbf{Avg} & \textbf{Last} \\
\midrule
DyTox\cite{p18}       & \multirow{1}{*}{50/class} & 62.94 & 60.18 \\
\midrule
SimpleCIL\cite{p8}   & \multirow{7}{*}{0/class}  & 42.95 & 44.08 \\
L2P\cite{p9}         &                           & 50.45 & 48.72 \\
DualPrompt\cite{p10}  &                           & 52.28 & 50.46 \\
CODA-Prompt\cite{p11} &                           & 59.85 & 59.99 \\
S-iPrompt\cite{p12}   &                           & 61.16 & 60.46 \\
DUCT\cite{p13}        &                           & 67.16 & 67.01 \\
PINA\cite{p14}        &                           & 69.06 & 68.36 \\
\midrule
\textbf{Ours} & \textbf{0/class} & \textbf{71.12} & \textbf{69.97} \\
\bottomrule
\end{tabularx}
\vspace{-8pt} 
\end{table}

\begin{table}[t]
\centering
\caption{Ablation study of different components.}
\label{tab4}
\renewcommand{\arraystretch}{1.15}
\begin{tabularx}{\columnwidth}{c c c C C C}
\toprule
\textbf{Disen} & \textbf{Anti-for} & \textbf{Dom-inv} & \textbf{CDDB} & \textbf{CoRe50} & \textbf{DomainNet} \\
\midrule

\textcolor{red}{\ding{55}} & \textcolor{red}{\ding{55}} & \textcolor{red}{\ding{55}} & 81.21 & 86.13 & 60.34 \\

\textcolor{green}{\checkmark} & \textcolor{red}{\ding{55}} & \textcolor{red}{\ding{55}} & 84.53 & 87.86 & 68.59 \\
\textcolor{green}{\checkmark} & \textcolor{green}{\checkmark} & \textcolor{red}{\ding{55}} & 85.12 & 91.08 & 69.92 \\
\textcolor{green}{\checkmark} & \textcolor{green}{\checkmark} & \textcolor{green}{\checkmark} & \textbf{86.35} & \textbf{92.36} & \textbf{71.12} \\
\bottomrule
\end{tabularx}
\vspace{-18pt} 
\end{table}

\subsection{Ablation studies}
\subsubsection{Ablation on Module}
The first row in Table~\ref{tab4} shows the performance of CLIP on three datasets without introducing any decoupling representation modules. After introducing the decoupling representation module, the average accuracy of the model on the three datasets increased by 3.32\%, 1.73\%, and 8.25\% respectively. This substantial performance leap verifies that in domain incremental learning, stripping away non-essential domain-specific noise is crucial for extracting robust category features. On this basis, the addition of the weight fusion mechanism further improved the performance by 0.59\%, 3.22\%, and 1.33\%. This indicates that the weight fusion mechanism effectively consolidates historical memory while ensuring the plasticity of the model. In addition, the feature augmentation strategy improves the model performance by 1.23\%, 1.28\%, and 1.2\%. The complete method that integrates all modules achieves the best performance.

\subsubsection{Sensitivity Analysis on q and $\lambda$}
\begin{figure}[]
    \centering
    \begin{subfigure}{0.49\linewidth}
        \centering
        \includegraphics[width=\linewidth]{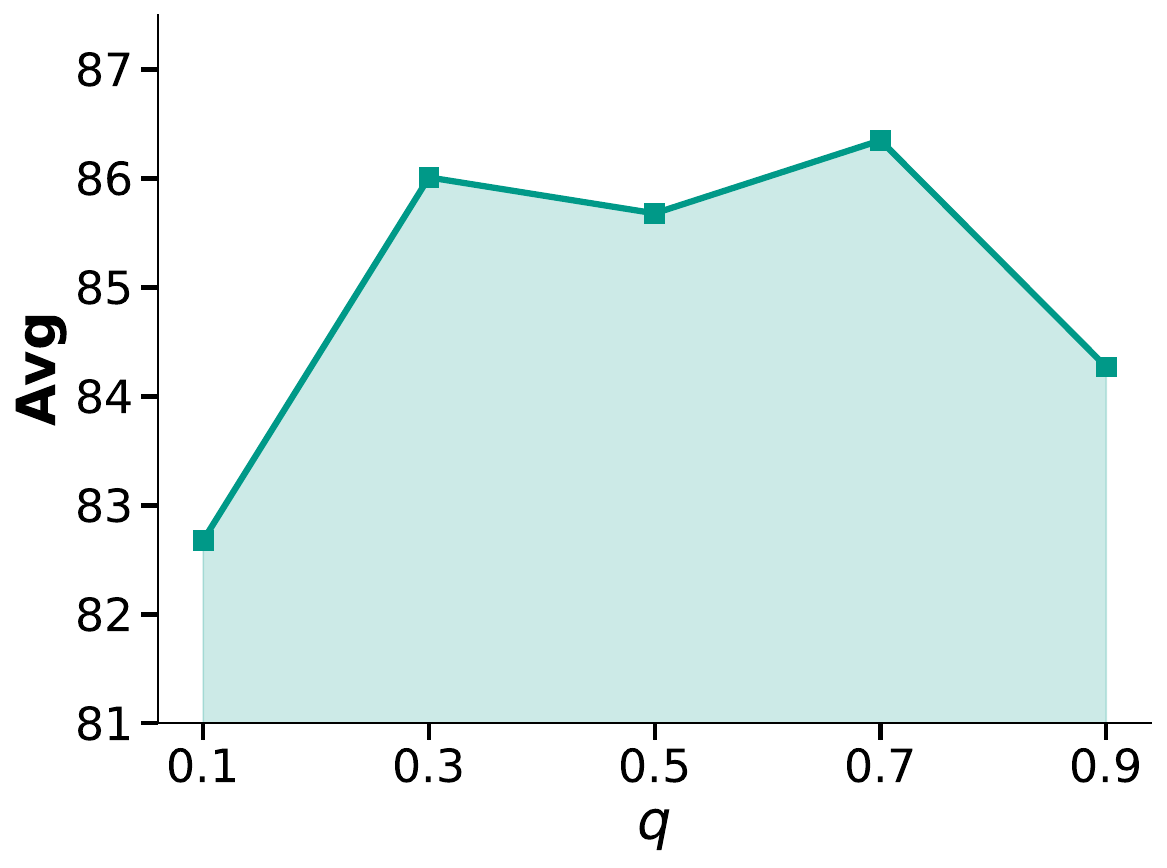} %
        \label{fig3:left}
    \end{subfigure}
    \hfill
    \begin{subfigure}{0.49\linewidth}
        \centering
        \includegraphics[width=\linewidth]{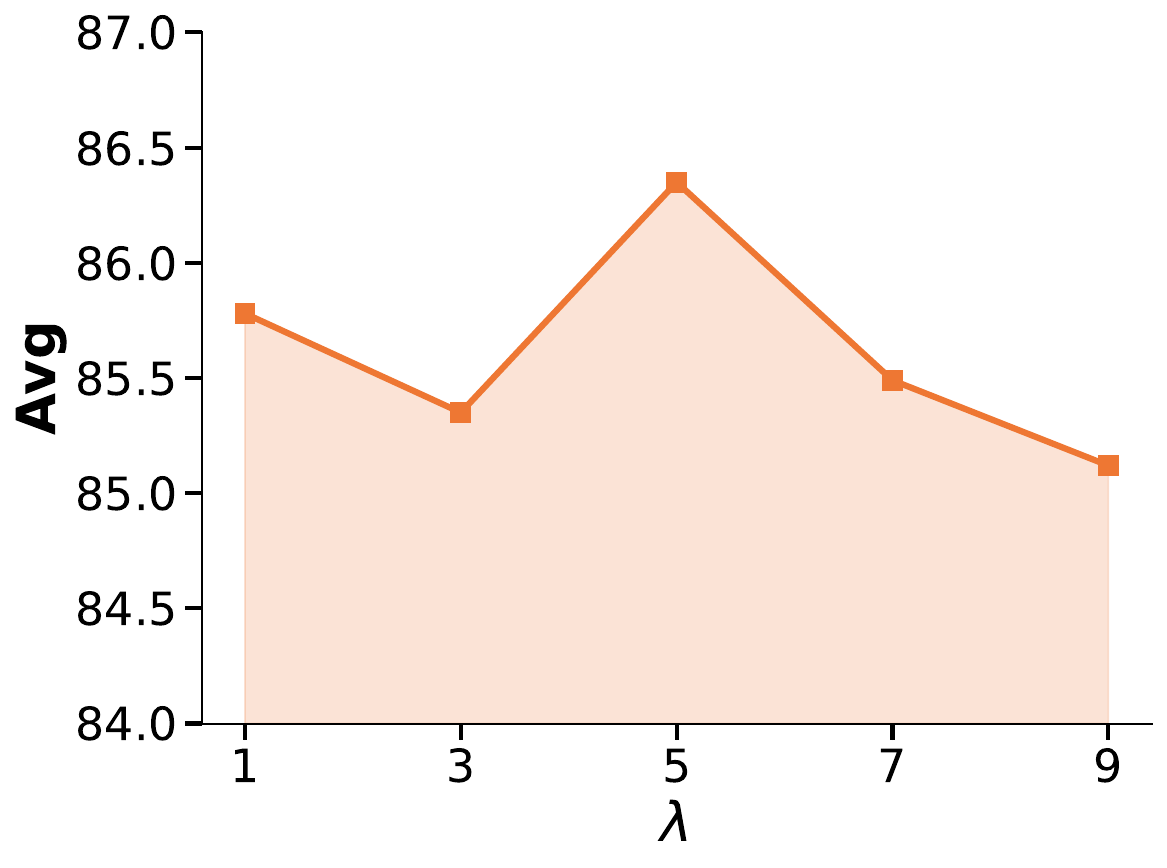} %
        \label{fig3:right}
    \end{subfigure}
    \caption{The sensitivity analysis on $q$ and $\lambda$ with CDDB}
    \label{fig3}
\vspace{-21pt}
 
\end{figure}

We conducted a sensitivity analysis on the hyperparameters $q$ representing the amplification degree in gce loss and $\lambda$ denoting the weight for feature augmentation as shown in Fig.~\ref{fig3}. By varying $q \in \{0.1, 0.3,0.5, 0.9\}$ with fixed $\lambda=5$, and $\lambda \in \{1, 3,5,7, 9\}$ with fixed $q=0.7$, we observed that the model yields stable and peak performance at the configuration of $q=0.7$ and $\lambda=5$.

\section{Conclusion}
This paper proposes a domain incremental learning framework that does not require examples and is domain-independent. In response to the limitations of existing methods, which rely on domain ids and have insufficient generalization, we effectively disentangle the representation mechanism to remove non-essential style noise and extract intrinsic features that are shared across domains. At the same time, we utilize the weight fusion strategy to dynamically integrate new and old knowledge in the parameter space, significantly reducing catastrophic forgetting without storing old samples. Future work will focus on complex scenarios involving both classes and domains, expand the framework to distinguish between style shifts and semantic novelty, and construct a lifelong learning system that is adaptable to the open world.

\bibliographystyle{IEEEbib}
\bibliography{refs}

\end{document}